\documentclass[10pt,twocolumn,letterpaper]{article}

\usepackage[pagenumbers]{cvpr} % To force page numbers, e.g. for an arXiv version

\usepackage[normalem]{ulem}  % normalem preserves the normal \emph{} behavior
\usepackage{xcolor} % For color support

\usepackage{graphicx}
\usepackage{amsmath}
\usepackage{amssymb}
\usepackage{booktabs}
\usepackage{blindtext}
\usepackage{multirow}
\usepackage{makecell}
\usepackage{colortbl}
\usepackage{adjustbox}
\usepackage{booktabs}    % For better table lines
\usepackage{multirow}    % For merged cells
\usepackage{threeparttable}  % For table notes
\usepackage{amssymb}   
\usepackage{wasysym}  % For circle symbols

\usepackage{pifont}     % For dingbats (stars)
\usepackage{tikz}       % For bar representation

\usepackage{pifont}     % For dingbats (stars)
\usepackage{tikz}       % For bar representation

\definecolor{dangerred}{RGB}{211,47,47}

\definecolor{successgreen}{RGB}{20,140,120}   % success / high
\definecolor{dangerred}{RGB}{200,55,65}

\definecolor{cvprblue}{rgb}{0.21,0.49,0.74}
\usepackage[pagebackref,breaklinks,colorlinks,allcolors=cvprblue]{hyperref}

\usepackage{mathtools, float, graphicx, caption, multirow}
\usepackage{algorithm, algpseudocode}
\usepackage{tcolorbox}
\usepackage{balance}

\definecolor{myblue}{rgb}{0.9,0.95,1.0}
\definecolor{mygreen}{rgb}{0.9,1.0,0.9}

\usepackage[accsupp]{axessibility}

\usepackage{sidecap}
\usepackage{colortbl} % For table row colors
\definecolor{lightgray}{RGB}{240,240,240} % Define a subtle gray for alternating rows
\definecolor{headerblue}{RGB}{200,220,255} % Define a soft blue for headers
\definecolor{headeryellow}{RGB}{255, 255, 180} % Soft yellow for headers
\definecolor{headerlavender}{RGB}{200, 200, 255} % Lavender-like pastel for headers

\definecolor{cvprblue}{rgb}{0.21,0.49,0.85}
\definecolor{lightblue}{RGB}{100,150,255}
\hypersetup{colorlinks=true, linkcolor=red, citecolor=lightblue, urlcolor=blue}
\def\paperID{} % *** Enter the Paper ID here
\def\confName{CVPR}
\def\confYear{2026}

\title{\texttt{RAVEN}: Erasing Invisible Watermarks via Novel View Synthesis}

\author{Fahad Shamshad$^{1}$ \quad Nils Lukas$^{1}$ \quad Karthik Nandakumar$^{1,2}$\\
\newline
$^{1}$MBZUAI, UAE \quad
$^{2}$Michigan State University, USA\\
{\tt\small \{fahad.shamshad, nils.lukas, karthik.nandakumar\}@mbzuai.ac.ae}
}

\begin{document}
\iffalse
\twocolumn[{%
\renewcommand\twocolumn[1][]{#1}%
\maketitle%

\begin{center}
    \centering
    \includegraphics[width=\textwidth, trim = 0.5cm 2.1cm 3.2cm 0.8cm, clip]{figures/teaser_1.pdf}
    \captionof{figure}{}
    \label{fig:introduction}
\end{center}
}]
\fi
\maketitle
\begin{abstract}
Invisible watermarking has become a critical mechanism for authenticating AI-generated image content, with major platforms deploying watermarking schemes at scale. However, evaluating the vulnerability of these schemes against sophisticated removal attacks remains essential to assess their reliability and guide robust design. In this work, we expose a fundamental vulnerability in invisible watermarks by reformulating watermark removal as a view synthesis problem. Our key insight is that generating a perceptually consistent alternative ``view" of the same semantic content, akin to re-observing a scene from a shifted perspective, naturally removes the embedded watermark while preserving visual fidelity. This reveals a critical gap: watermarks robust to pixel-space and frequency-domain attacks remain vulnerable to semantic-preserving viewpoint transformations. We introduce a zero-shot diffusion-based framework that applies controlled geometric transformations in latent space, augmented with view-guided correspondence attention to maintain structural consistency during reconstruction. Operating on frozen pre-trained models without detector access or watermark knowledge, our method achieves state-of-the-art watermark suppression across 15 watermarking methods--outperforming 14 baseline attacks while maintaining superior perceptual quality across multiple datasets. Our code will be made publicly available.
\end{abstract}

\section{Introduction}
\label{sec:intro}

Invisible content watermarking has emerged as a key solution for tracking the provenance of AI-generated images~\cite{zhao2411sok,vsarvcevic2024u,xu2024copyrightmeter}. These techniques embed imperceptible patterns that remain reliably detectable with a secret key even after common post-processing operations, a property typically referred to as \textit{robustness}. 
Regulatory frameworks such as the EU AI Act~\cite{rijsbosch2025adoption} and the U.S. Executive
Order on AI~\cite{biden2023executive} now explicitly call for watermarking of AI-generated
content to help prevent misuse and support authenticity verification.
Major generative systems (e.g., SynthID from Google~\cite{gowal2025synthid}) now incorporate watermarking into hundreds of millions of images. As the need for trusted AI-generated content continues to grow, evaluating the vulnerability of watermarking methods to removal attacks becomes critical to assess the usefulness of deployed systems and guide the development of more robust watermarking schemes~\cite{anwaves,diaa2024optimizing}.

\begin{table}[t]
\centering
\caption{Comparison of watermark removal methods across three dimensions: (i) \textbf{Effectiveness}, \textit{i.e.}, how strongly the attack suppresses both pixel-level and semantic watermarks; (ii) \textbf{Quality}, preservation of semantic content and visual naturalness; and (iii) \textbf{Efficiency}, compute budget in training and per-image processing. 
Regen~\cite{zhao2024invisible} is efficient but limited to weak watermarks; CtrlGen+~\cite{liu2024image} requires multi-node training (8 GPUs); IRA~\cite{muller2025black} performs slow per-image optimization ($\approx40$ min./image) and require access to model parameter to be effective; and UnMarker~\cite{kassis2025unmarker} degrades visual fidelity. Our method achieves superior performance across all dimensions.}
\label{tab:comparison_approaches}
% Define colors

\resizebox{0.96\linewidth}{!}{
\begin{tabular}{l*{3}{c}}
\toprule 
 \multirow{1}{*}{\textbf{Approach}} & 
\multirow{1}{*}{\textbf{Effective}} & 
\multicolumn{1}{c}{\textbf{Quality}} & 
\multirow{1}{*}{\textbf{Efficiency}} \\
\midrule
$\text{Regen}_{\text{\color{blue}(NeurIPS'24)}}$~\cite{zhao2024invisible}        & {$\color{successgreen}\CIRCLE\CIRCLE\color{dangerred}\CIRCLE\CIRCLE$} & {$\color{successgreen}\CIRCLE\CIRCLE\color{dangerred}\CIRCLE\CIRCLE$} & {\color{successgreen}$\CIRCLE\CIRCLE\CIRCLE\CIRCLE$}  \\
$\text{CtrlGen+}_{\text{\color{blue}(ICLR'25)}}$~\cite{liu2024image}        & {$\color{successgreen}\CIRCLE\CIRCLE\CIRCLE\color{dangerred}\CIRCLE$} & {$\color{successgreen}\CIRCLE\CIRCLE\CIRCLE\color{dangerred}\CIRCLE$} & {\color{dangerred}$\CIRCLE\CIRCLE\CIRCLE\CIRCLE$}  \\
$\text{IRA}_{\text{\color{blue}(CVPR'25)}}$~\cite{muller2025black}     & {$\color{successgreen}\CIRCLE\CIRCLE\color{dangerred}\CIRCLE\CIRCLE$} & {$\color{successgreen}\CIRCLE\CIRCLE\CIRCLE\color{dangerred}\CIRCLE$} & {\color{dangerred}$\CIRCLE\CIRCLE\CIRCLE\CIRCLE$} \\
$\text{UnMarker}_{\text{\color{blue}(S\&P'25)}}$~\cite{kassis2025unmarker}  & {\color{successgreen}$\CIRCLE\CIRCLE\CIRCLE\CIRCLE$} & {$\color{successgreen}\CIRCLE\CIRCLE\CIRCLE\color{dangerred}\CIRCLE$} & {$\color{successgreen}\CIRCLE\color{dangerred}\CIRCLE\CIRCLE\CIRCLE$}  \\
\midrule
\rowcolor{gray!20}\textbf{\texttt{RAVEN}} (Ours) & {\color{successgreen}$\CIRCLE\CIRCLE\CIRCLE\CIRCLE$} & {\color{successgreen}$\CIRCLE\CIRCLE\CIRCLE\CIRCLE$} & {\color{successgreen}$\CIRCLE\CIRCLE\CIRCLE\CIRCLE$} \\
\bottomrule
\end{tabular}
}

\begin{tablenotes}
\item[] \hspace{0em} \small {\color{successgreen}$\CIRCLE\CIRCLE\CIRCLE\CIRCLE$} High \hspace{2em} {$\color{successgreen}\CIRCLE\CIRCLE\color{dangerred}\CIRCLE\CIRCLE$} Moderate \hspace{2em} {\color{dangerred}$\CIRCLE\CIRCLE\CIRCLE\CIRCLE$} Low
\end{tablenotes}
\end{table}
%%%%%%%%%%%%%%%%%%%%%%%%%%%%%%%%%%%%%%%%%%%%%%%%%%%%%%%%%%%%%%%%

An ideal watermark removal attack must strike the right balance between perceptual fidelity and watermark suppression~\cite{zhao2024sok,holtervennhoff2025security,lukas2023ptw}. Perceptual fidelity refers to preserving the semantic content, fine-grained textures, and the absence of visible artifacts detectable by human observers or automated quality metrics. On the other hand, watermark suppression requires that the transformed 
image reliably evades detection, rendering the embedded signal 
statistically indiscernible to the watermark detector.  A successful attack produces images that are perceptually identical 
to the original, while successfully defeating the detection mechanism~\cite{ding2024erasing}.

Existing watermark removal attacks operate either in pixel space using signal-processing operations or in latent space through diffusion-based purification~\cite{zhao2411sok,wen2023tree}. However, both families struggle to jointly preserve perceptual fidelity and suppress watermark detectability (Tab.~\ref{tab:comparison_approaches}). Pixel-space methods, such as compression, filtering, and noise injection, often fail against modern semantic watermarking schemes and can introduce visible artifacts~\cite{an2024waves}. Latent-space diffusion approaches can suppress stronger watermarks, but they typically rely on substantial noise injection, which disrupts scene structure and degrades visual realism~\cite{zhao2024invisible}. In addition, many state-of-the-art methods incur high computational cost~\cite{liu2024image}, requiring several minutes per image~\cite{muller2025black}, or depend on privileged information such as access to the watermark decoder~\cite{lukas2023leveraging} or watermarked training data~\cite{saberi2023robustness}, which limits their practical applicability.

In this work, we expose a fundamental vulnerability in invisible watermarking by re-conceptualizing removal as a novel view synthesis problem~\cite{watson2022novel}. Our key insight is that generating a perceptually consistent alternative ``view’’ of the same semantic content produces a new image instance that preserves visual fidelity while statistically decorrelating from the embedded watermark signal. This exposes a critical gap in current robustness evaluations: even watermarks that can withstand pixel-space manipulation and latent-space purification remain vulnerable to semantic-preserving viewpoint transformations. Crucially, our approach does not require access to watermark detectors, no knowledge of the watermarking scheme, and no additional data or retraining, allowing deployment in practical settings. 

To realize our view synthesis attack, we introduce a zero-shot diffusion-based framework that modulates latent representations to simulate novel viewpoints while preserving semantic content. Our method partially inverts the watermarked image into the diffusion latent space, exposing the watermark-entangled representation while preserving the semantic structure of the scene. We then apply subtle latent-space viewpoint shifts that introduce geometric variations sufficient to disrupt watermark alignment without altering semantic image content. To maintain high perceptual fidelity during reconstruction, we replace conventional self-attention with a view-guided correspondence attention mechanism that enforces structural consistency between the original and synthesized views. Notably, our approach operates in a zero-shot manner using a frozen pre-trained image-to-image diffusion model and requires only the watermarked image as input, enabling deployment across diverse watermarking schemes without adaptation or retraining. Our contributions are as follows:

\begin{itemize}
    \item \textbf{Novel Attack Vector via View Synthesis}: We expose a critical gap in watermark robustness evaluation by reframing removal as a view synthesis problem: synthesizing a perceptually consistent alternative view of the same scene preserves semantics while statistically decorrelating from the original watermark embedding.

    \item \textbf{Latent Viewpoint Modulation Framework}: We design a zero-shot diffusion-based pipeline that removes watermarks by applying controlled viewpoint shifts in latent space, producing perceptually consistent images without requiring access to watermark decoder or knowledge of the watermarking scheme.

    \item \textbf{View-Guided Correspondence Attention}: We introduce a correspondence-aware attention mechanism that maintains structural and appearance alignment between the original and synthesized views, enabling watermark suppression without altering fine-grained visual details.

    \item \textbf{Extensive Evaluation}: We demonstrate state-of-the-art watermark removal across multiple watermarking schemes and evaluate robustness under various attack conditions. Our approach attains superior watermark suppression with minimal perceptual degradation.

\end{itemize}

\section{Related Work}

\noindent \textbf{Image Watermarking.} Image watermarking enables provenance verification of AI-generated content by embedding imperceptible signals that can later be decoded for authentication~\cite{zhao2411sok,vsarvcevic2024u,xu2024copyrightmeter}. Existing techniques can be grouped by when the watermark is inserted and where it resides in the representation. Post-hoc methods operate directly on generated images using signal processing or encoder-decoder architectures (\textit{e.g.}, DwtDCT~\cite{cox2008digital}, RivaGAN~\cite{zhang2019robust}, StegaStamp~\cite{tancik2020stegastamp}), offering broad applicability but often degrading under denoising, compression, or cropping. In contrast, in-generation watermarking modifies the generative model or its latent inputs, such as VAE-decoder tuning (StableSignature~\cite{fernandez2023stable}), latent distribution shaping (Gaussian Shading~\cite{yang2024gaussian}), or noise-pattern encoding (TreeRing~\cite{wen2023tree}) to achieve stronger robustness by coupling the watermark with semantic structure. While these schemes are primarily evaluated for detection accuracy and robustness against distortions, their resilience to sophisticated removal attacks remains an active area of investigation~\cite{anwaves}.

\noindent \textbf{Watermark Removal Attacks.} Efforts to remove embedded watermarks can be characterized by their threat model assumptions and attack primitives. \textit{Signal processing methods} apply standard image transformations such as JPEG compression, geometric distortions, cropping, and additive noise~\cite{anwaves}. While computationally efficient and requiring no model access, these approaches offer limited effectiveness against modern robust watermarking schemes~\cite{ding2024waves}. \textit{Regeneration attacks} leverage diffusion models~\cite{zhao2024invisible} or VAEs~\cite{lukas2023leveraging} to reconstruct images by injecting and removing noise, effectively eliminating pixel-level watermarks but often fail against  semantic watermarks. Moreover, achieving removal of stronger watermarks requires substantial noise injection, which rapidly degrades structural consistency and visual realism, limiting practical applicability. Adversarial optimization-based methods treat the detector as a classifier and craft perturbations to induce detection failure~\cite{zhao2024sok}, but typically rely on unrealistic capabilities—white-box access~\cite{lukas2023leveraging} or surrogate training using non-watermarked data~\cite{saberi2023robustness,wu2023tune}. Existing approaches thus exhibit a persistent 
limitation: black-box methods struggle with semantic watermarks or quality preservation, while effective removal presumes impractical attacker capabilities.

\noindent \textbf{Diffusion Models in Vision.} Diffusion models have emerged 
as powerful generative frameworks for image synthesis and editing~\cite{huang2025diffusion,wu2023tune}. Recent advances in conditioning mechanisms and latent space manipulation enable semantic-preserving transformations while modifying 
low-level features. Diffusion models also excel at view synthesis and 
appearance editing through controlled latent modulation. However, these NVS approaches 
generally require dedicated training on task-specific datasets with multi-view 
supervision~\cite{chan2023generative,yu2024viewcrafter}. Recent zero-shot image-to-video frameworks demonstrate that pretrained diffusion models can maintain semantic and structural consistency across transformed latent representations, even without retraining or paired data~\cite{khachatryan2023text2video}. Motivated by this observation, we investigate whether similar latent viewpoint transformations can be leveraged for watermark removal, aiming to decorrelate embedded watermark signals while preserving perceptual identity. \textit{To the best of our knowledge, we are the first to explore diffusion-based novel-view synthesis for invisible watermark removal}.

\section{Threat Model}
\label{sec:threat}

We formulate the watermark removal problem within a realistic
threat model involving two key entities: a \textbf{provider} who operates a watermark-protected image generation service, and an \textbf{attacker} attempting to erase watermark evidence while retaining the semantic utility of the image.

\noindent \textbf{Provider's Goals and Capabilities}: The provider watermarks the generated images to establish provenance and prevent misuse of generated content. They may adopt either post-hoc or in-generation watermarking schemes and possess full control over the deployed generator and watermark detector. Detection requires only a single image. We assume that the provider has sufficient computational resources and data to train a robust watermarking system.

\noindent \textbf{Attacker's Goals and Capabilities}: The attacker aims to remove watermarks while preserving semantic content, producing alternative observations of the same scene rather than exact pixel reconstructions. The attacker operates under realistic constraints across four dimensions: \textit{(i) \textbf{Access}}: no knowledge of model parameters, watermarking schemes, or detector internals; no access to detection APIs for iterative queries; \textit{(ii) \textbf{Data}}: no paired clean-watermarked examples, no proprietary datasets for surrogate training, and access to only a single input image; \textit{(iii) \textbf{Compute}}: limited resources for multi-GPU training or fine-tuning, limited to lightweight inference on consumer hardware; \textit{(iv) \textbf{Time}}: per-image processing must complete within seconds, precluding minutes-long optimization. The attacker may use publicly available pretrained image-to-image diffusion models in a zero-shot manner, but lacks the compute or data to fine-tune or retrain them for removal or surrogate-detector training. This setting reflects practical scenarios where adversaries must succeed without privileged access, extensive datasets, or prolonged computation.

\begin{figure*}[t]
\centering
\begin{minipage}[c]{0.69\linewidth}
    \centering
    \includegraphics[width=\linewidth,trim={5em 4em 20em 3em},clip]{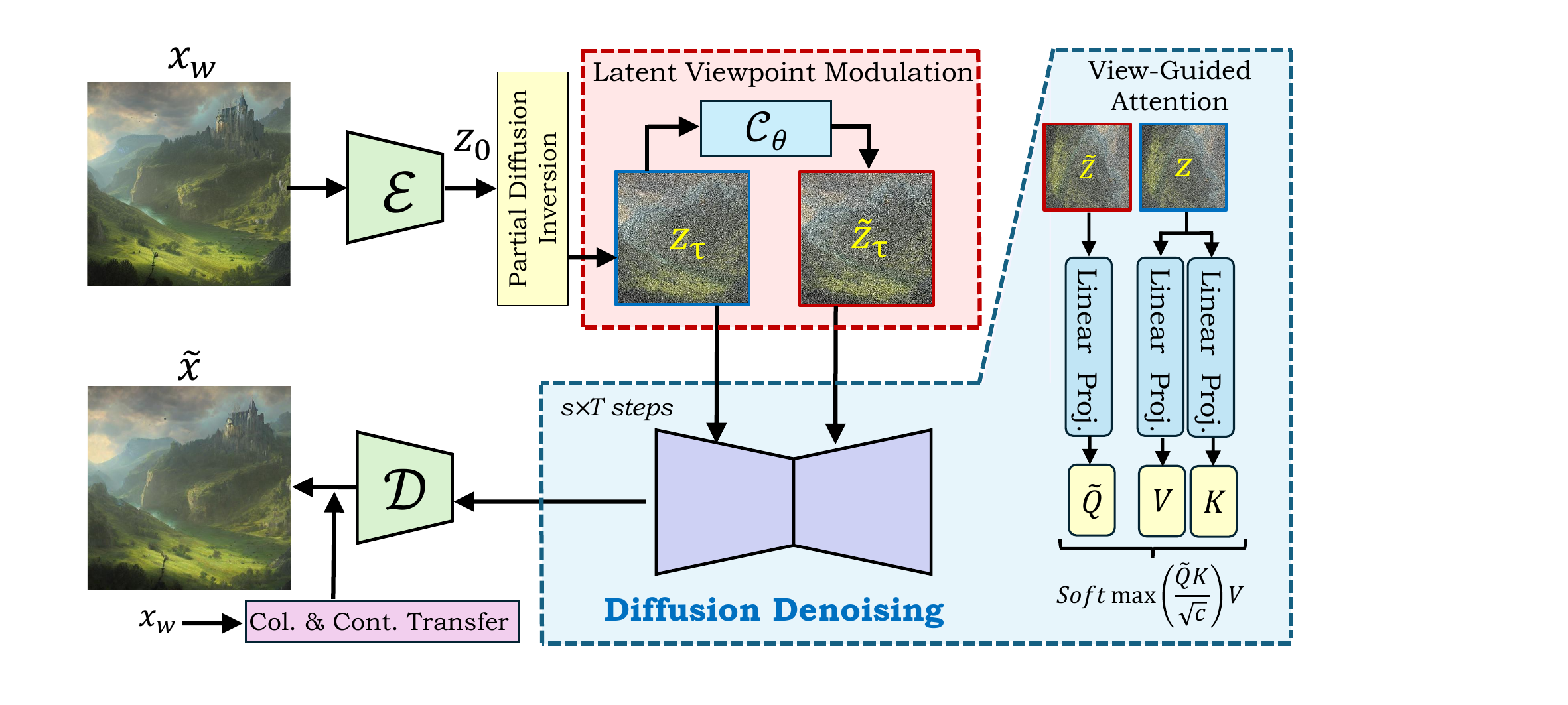}
\end{minipage}
\hfill
\begin{minipage}[c]{0.30\linewidth}
    \caption{\textbf{Overview of \texttt{RAVEN}.} Given a watermarked image $x_w$, we 
    encode it into latent space and perform partial diffusion inversion 
    to obtain $z_\tau$. Latent viewpoint modulation applies a camera motion $\mathcal{C}_\theta$ to generate $\tilde{z}_\tau$. During denoising 
    steps, view-guided correspondence attention computes queries 
    $\tilde{Q}$ from the transformed latent while keys $K$ and values $V$ 
    come from a reference latent denoised in parallel from $z_\tau$. 
    This cross-view mechanism preserves scene semantics while suppressing 
    watermark-related cues. The decoder $\mathcal{D}$ reconstructs the 
    output, followed by color and contrast correction to produce the final 
    watermark-free result $\tilde{x}$.}
    \label{fig:main_diagram}
\end{minipage}
\end{figure*}

\section{Proposed Approach}

\subsection{Background and Problem Statement}
\label{sec:prel}

\noindent \textbf{Image-to-Image Diffusion Models}:
We build our method on latent image-to-image diffusion models, instantiated with Stable Diffusion~\cite{rombach2022high}, which operate in the latent space of a pre-trained autoencoder $(\mathcal{E}, \mathcal{D})$. Let $x \in \mathcal{X}$ be an image sampled from the image space $\mathcal{X}$, $z_0 = \mathcal{E}(x) \in \mathcal{Z}$ be the corresponding latent from the latent space $\mathcal{Z}$ and $x' =  \mathcal{D}(z_0) \approx x$. The forward process adds noise to latent $z_0$ via $q(z_t | z_0) = \mathcal{N}(z_t; \sqrt{\bar{\alpha}_t}z_0, (1-\bar{\alpha}_t)\mathbf{I})$, while a learned denoiser $\epsilon_\theta(z_t, t)$ reverses this process to recover $z_0$. Unlike text-to-image generation from random noise, image-to-image diffusion inverts a real image into noisy latent $z_\tau$ via DDIM inversion - optionally in a partial timestep $\tau < T$ to preserve the semantic structure - then applies transformations $\mathcal{T}: \mathcal{Z} \to \mathcal{Z}$ in latent space before reconstructing $x' = \mathcal{D}(\text{DDIM-Sample}(z'_\tau))$. This latent-space control enables image-to-image transformations that maintain semantic content while altering representational factors, making such models well suited for watermark removal attacks.

\noindent \textbf{Watermarking System}:
We model a watermarking system as 
$\mathcal{W} = (\mathcal{U}_\zeta, \mathcal{V}_\eta)$, where encoder 
$\mathcal{U}_\zeta: \mathcal{X} \times \mathcal{K} \to \mathcal{X}$ embeds watermark key $\kappa \in \mathcal{K}$ into image $x$ to produce watermarked image $x_w = \mathcal{U}_\zeta(x, \kappa)$, and detector $\mathcal{V}_\eta: \mathcal{X} \to \mathcal{K} \cup \{\bot\}$ extracts the embedded key $\hat{\kappa} = 
\mathcal{V}_\eta(x_w)$ or returns $\bot$ if no watermark is detected. Modern watermarks operate at two levels: \textit{pixel-level} schemes embed signals in spatial or frequency domains (e.g., DwtDct, StegaStamp), while \textit{semantic-level} schemes couple signals with latent structure in generative models (e.g., Tree-Ring, StableSignature) for stronger robustness. An effective watermark satisfies \textit{imperceptibility} ($x_w \approx x$ perceptually) and \textit{robustness} (invalidating $\kappa$ requires noticeable degradation). In realistic deployments, both $\mathcal{U}_\zeta$ and $\mathcal{V}_\eta$ are proprietary, exposed only through black-box APIs.

\noindent \textbf{Problem Formulation.}
Given a watermarked image $x_w = \mathcal{U}_\zeta(x,\kappa)$ returned by a provider’s generative API, our goal is to design a transformation $\mathcal{A}:\mathcal{X}\!\to\!\mathcal{X}$ producing $\tilde{x} = \mathcal{A}(x_w)$ that removes the watermark (i.e., $\mathcal{V}_\eta(\tilde{x}) = \bot$ with high probability) while preserving the content (i.e., $\tilde{x} \approx x_w$). 
The attacker operates under strict no-box constraints (Sec. \ref{sec:threat}): no 
access to $(\mathcal{U}_\zeta,\mathcal{V}_\eta,\zeta,\eta,\kappa)$, no detector 
queries, no clean-watermark paired image supervision, but may use publicly available image-to-image diffusion models.
Let $d: \mathcal{K} \times \mathcal{K} \to 
\mathbb{R}$ be an unknown distance metric in the key space $\mathcal{K}$. A successful attack must satisfy:
\begin{align}
\textbf{(P1) Detection Evasion:} \quad & d(\mathcal{V}_\eta(\tilde{x}),\kappa) > \phi \\
\textbf{(P2) Semantic Preservation:} \quad & \mathcal{S}((\tilde{x}),x_w) \text{ is high} \\
\textbf{(P3) Visual Naturalness:} \quad & \mathcal{N}(\tilde{x}) \text{ is high}
\end{align}
where $\phi$ is the detection threshold, $\mathcal{S}$ denotes semantic similarity (e.g., CLIP-based), and $\mathcal{N}$ measures naturalness. We assume that the detector outputs $\bot$ when $d(\mathcal{V}_\eta(\cdot),\kappa) > \phi$, but the attacker does not know either $d$ or $\phi$. The removal problem thus requires jointly maximizing detection evasion (\textbf{P1}) while maintaining semantic consistency (\textbf{P2}) and naturalness (\textbf{P3}), objectives that are inherently coupled and often conflicting: stronger watermark suppression tends to degrade semantic or visual fidelity, while conservative edits may fail to evade detection. Striking the right balance between removal efficacy and content preservation is critical to effective removal of watermarks.

\subsection{Our Method: \textbf{\large{\texttt{RAVEN}}}}

Our goal is to remove invisible watermarks while preserving the semantic and perceptual integrity of the image. We approach this challenge by reframing watermark removal as a NVS problem, generating alternative observations of the same scene that remain semantically consistent but disrupt the spatial alignment and statistical coupling of the embedded watermark signal. This perspective departs from traditional signal-processing or advanced diffusion-based watermark removal methods, shifting the focus from erasing artifacts to reconstructing a perceptually consistent alternative view that, by virtue of its altered viewpoint, breaks the spatial and representational dependencies exploited by watermark encoders. 
%%%%%%%%%%%%%%%%%%%%%%%%%%%%%%%%%%%%%%%%%%%%%%%%%%%%%
A typical way to achieve this goal is to use recent advanced NVS pipelines, most of which are built on diffusion models trained for 3D-aware generation or video synthesis~\cite{chan2023generative,yu2024viewcrafter}.  However, such frameworks rely on extensive multi-view training data and high computational resources, making them impractical for watermark removal where paired supervision and clean references are often unavailable.

\noindent \textbf{Overview of Proposed Solution}: To overcome these limitations, we propose \texttt{RAVEN}, a zero-shot diffusion-based framework designed to remove watermarks without requiring access to a watermark detectors, embedding keys, or paired clean-watermark images training data. Our key insight is that pre-trained image-to-image diffusion models already encode rich geometric and semantic priors sufficient for view synthesis—they simply require lightweight modifications to maintain consistency across synthesized views. 
As illustrated in Fig.~\ref{fig:main_diagram}, \texttt{RAVEN} operates in three stages: (i)~\textit{\textbf{partial diffusion inversion}}, where we encode the watermarked image into a noisy latent representation, preserving semantic structure while exposing the watermark-entangled representation; (ii)~\textit{\textbf{latent viewpoint modulation}}, where we apply controlled geometric transformations to simulate viewpoint shifts that disrupt watermark alignment; and (iii)~\textit{\textbf{view-guided correspondence attention}}, which replaces standard self-attention with cross-frame attention mechanisms during the denoising process to preserve structural consistency.  The following sections describe these stages in detail.

\subsubsection{Partial Diffusion Inversion}

Given a watermarked image $x_w$, we first map it into the diffusion model’s latent space using the encoder $\mathcal{E}(\cdot)$, producing a latent code $z = \mathcal{E}(x_w)$. We apply partial diffusion inversion to introduce controlled noise into the latent while keeping its overall structure intact. To achieve this, we add noise corresponding to an intermediate diffusion timestep $\tau = \lfloor s \cdot T \rfloor$, where $T$ is the total number of diffusion steps and $s \in [0,1]$ controls the noise level. The forward diffusion update is:

\begin{equation}
z_\tau = \sqrt{\bar{\alpha}_\tau}\, z \;+\; \sqrt{1 - \bar{\alpha}_\tau}\, \epsilon, 
\qquad \epsilon \sim \mathcal{N}(0, I),
\end{equation}

\noindent where $\bar{\alpha}_\tau = \prod_{t=1}^{\tau} \alpha_t$ denotes the cumulative noise schedule.
The parameter $s$ controls the trade-off between noise strength and semantic preservation: 
small values of $s$ retain finer details, whereas large values introduce more stochasticity 
at the cost of potential semantic drift. The resulting latent $z_\tau$ maintains the overall 
scene structure while injecting the controlled randomness required for the subsequent 
viewpoint-modulation stage.

\subsubsection{Latent Viewpoint Modulation}

Given the partially inverted latent $z_\tau$, our goal is to introduce a small, semantically consistent viewpoint change directly in latent space. Such a shift acts like a subtle camera motion: it displaces scene structures coherently while preserving the overall layout. We model this operation using a spatial warping function 
$\mathcal{C}_\theta : \mathbb{R}^2 \rightarrow \mathbb{R}^2$, which specifies 
the sampling location for each latent feature. The modulated latent 
$\tilde{z}_\tau$ is obtained as:

\begin{equation}
\tilde{z}_\tau[i,j] = z_\tau[\mathcal{C}_\theta(i,j)].
\end{equation}

\noindent Here, $\tilde{z}_\tau$ is a geometrically transformed version of $z_\tau$, where 
each spatial location $(i,j)$ in the output latent fetches its content from a 
shifted location determined by $\mathcal{C}_\theta(i,j)$. This warping 
operation induces a coherent displacement of scene structures in the latent 
space, space—analogous to observing the same scene from a slightly shifted camera 
position, while preserving the global semantics encoded in $z_\tau$.

Conventional NVS pipelines handle large viewpoint changes and therefore rely on multi-view training data, depth estimation, and heavy 3D consistency modeling~\cite{chan2023generative,yu2024viewcrafter}. Such complexity is unnecessary for our goal. We do not seek to reconstruct unseen geometry, but only to introduce a small, coherent perturbation in viewpoint that slightly alters spatial correspondence while preserving semantic content.  
For example, a simple and effective choice is a \textit{global camera translation},  
$\mathcal{C}_\theta(i,j) = (i + \Delta_x,\; j + \Delta_y)$,  
where $(\Delta_x, \Delta_y)$ determines the direction and magnitude of the apparent camera motion. This transformation induces a subtle yet spatially coherent shift in scene structure while maintaining semantic stability. Although more general motions are possible, small global translations strike the best balance between simplicity and consistency in our setting.

\definecolor{Gray}{gray}{0.93}
\begin{table*}[t]
\centering
\scriptsize
\caption{Verification performance of different watermarking methods under various attacks. \textbf{TPR@1\%FPR} is reported for in-generation semantic watermarking methods (TreeRing to ROBIN), where lower values indicate better attack performance. \textbf{Bit Accuracy} is reported for post-hoc bitstream-based methods (DwtDct to VINE), where values near 0.5 indicate successful watermark randomization. \texttt{RAVEN} achieves the lowest detection rates across both categories, demonstrating superior removal efficacy while maintaining visual quality.}
\begin{adjustbox}{width=\textwidth}
\begin{tabular}{l|>{\columncolor{myblue!50}}c>{\columncolor{myblue!50}}c>{\columncolor{myblue!50}}c>{\columncolor{myblue!50}}c>{\columncolor{myblue!50}}c>{\columncolor{myblue!50}}c|c|>{\columncolor{mygreen!50}}c>{\columncolor{mygreen!50}}c>{\columncolor{mygreen!50}}c>{\columncolor{mygreen!50}}c>{\columncolor{mygreen!50}}c>{\columncolor{mygreen!50}}c>{\columncolor{mygreen!50}}c>{\columncolor{mygreen!50}}c|c}
\toprule
\small \rotatebox{70}{Attack} & \rotatebox{70}{TreeRing~\cite{wen2023tree}} & \rotatebox{70}{Zodiac~\cite{zhang2024robust}} & \rotatebox{70}{HSTR~\cite{lee2025semantic}} & \rotatebox{70}{RingID~\cite{ci2024ringid}} & \rotatebox{70}{\shortstack{HSQR~\cite{lee2025semantic}\\\textcolor{blue}{\scriptsize }}} & \rotatebox{70}{ROBIN~\cite{huang2024robin}} & \rotatebox{70}{Avg.} & \rotatebox{70}{DwtDct~\cite{cox2008digital}} & \rotatebox{70}{DwtDctSvd~\cite{cox2008digital}} & \rotatebox{70}{RivaGAN~\cite{zhang2019robust}} & \rotatebox{70}{Stable Sign.~\cite{fernandez2023stable}} & \rotatebox{70}{Gauss Shad.~\cite{yang2024gaussian}} & \rotatebox{70}{TrustMark~\cite{bui2023trustmark}} & \rotatebox{70}{Stega St.~\cite{tancik2020stegastamp}} & \rotatebox{70}{VINE~\cite{lu2024robust}} & \rotatebox{70}{Avg.}\\
\midrule
\rowcolor{Gray} \multicolumn{17}{c}{\textbf{MS-COCO}~\cite{lin2014microsoft}} \\
\midrule
Clean    & 0.957 & 0.998 & 1.000 & 1.000 & 1.000 & 1.000 & 0.993 & 0.863 & 1.000 & 0.999 & 0.995 & 1.000 & 0.952 & 1.000& 1.000& 0.968  \\ 
\midrule
Bright.  & 0.463 & 0.843 & 0.899 & 0.988 & 0.991 & 0.985 & 0.862 & 0.572 & 0.555 & 0.862 & 0.894 & 0.962 & 0.918 & 0.992 & 0.998 & 0.839\\
Cont.    & 0.900 & 0.998 & 1.000 & 1.000 & 1.000 & 0.994 & 0.982 & 0.522 & 0.473 & 0.986 & 0.978 & 1.000 & 0.927 & 1.000  & 0.998 & 0.876\\
JPEG     & 0.548 & 0.973 & 0.994 & 1.000 & 1.000 & 0.986 & 0.917 & 0.516 & 0.602 & 0.821 & 0.806 & 0.992 &  0.919& 1.000 & 1.000 &0.851\\
Blur     & 0.934 & 0.998 & 1.000 & 1.000 & 1.000 & 0.991 & 0.987 & 0.677 & 1.000 & 0.998 & 0.911 & 1.000 &  0.942& 0.990 &0.973 &0.944\\
Noise    & 0.412 & 0.880 & 0.806 & 0.987 & 0.983 & 0.840 & 0.818 & 0.859 & 1.000 & 0.969 & 0.721 & 0.997 & 0.772 & 0.971 & 0.979 &0.878\\
BM3D     & 0.815 & 0.997 & 0.999 & 1.000 & 1.000 & 0.863 & 0.946 & 0.532 & 0.784 & 0.934 & 0.838 & 0.999 & 0.886 &0.997 & 1.000 & 0.886\\
\midrule
Center Crop     & 0.509 & 0.989 & 1.000 & 1.000 & 1.000 & 0.884 & 0.897 & 0.729 & 0.744 & 0.991 & 0.987 & 0.998 & 0.890 & 0.924& 0.953 & 0.890  \\
Random Crop     & 0.734 & 0.995 & 1.000 & 1.000 & 1.000 & 0.790 & 0.920 & 0.810 & 0.861 & 0.995 & 0.991 & 1.000 & 0.724 & 0.887 & 0.915 & 0.898\\
\midrule
VAE-B~\cite{balle2018variational}    & 0.509 & 0.944 & 0.973 & 0.992 & 0.992 & 0.817 & 0.871 & 0.523 & 0.648 & 0.570 & 0.717 & 0.982 &   0.834 & 0.883& 0.950 & 0.788\\
VAE-C~\cite{cheng2020learned}    & 0.536 & 0.958 & 0.982 & 1.000 & 1.000 & 0.820 & 0.883 & 0.521 & 0.596 & 0.552 & 0.715 & 0.987 & 0.847 & 0.870 & 0.915 & 0.778\\
Regen.~\cite{zhao2024invisible}   & 0.543 & 0.972 & 0.997 & 1.000 & 1.000 & 0.873 & 0.898 & 0.519 & 0.644 & 0.608 & \textbf{0.478} & 0.999 & 0.790 & 0.854 & 0.881 & 0.752\\
Rinse    & 0.461 & 0.905 & 0.982 & 0.997 & 0.999 & 0.793 & 0.856 & 0.508 & 0.561 & 0.527 & 0.494 & 0.999 & 0.713 & 0.809& 0.831 & 0.716\\
\midrule
CtrlGen+~\cite{liu2024image}   & 0.090 & 0.314 & 0.776 & 1.000 & 1.000 &0.328 & 0.585 & 0.532 & 0.517 & \textbf{0.481} & 0.564 & 1.000 & 0.683 & \textbf{0.557} & 0.887 & 0.640\\
UnMarker~\cite{kassis2025unmarker}     & 0.032 & 0.092 & 0.031&  0.257 & 0.018 &0.039 & 0.078 & \textbf{0.483} & 0.528 & 0.553 & 0.511 & 0.592 & 0.547 &0.636 & 0.617 & 0.573\\
\midrule
\texttt{{RAVEN}} (Ours)    & \textbf{0.020} & \textbf{0.067} & \textbf{0.025} & \textbf{0.018} & \textbf{0.015} & \textbf{0.012} & \textbf{0.026} & 0.517 & \textbf{0.472} & 0.499 & 0.519 &\textbf{0.540} & \textbf{0.476} & 0.575& \textbf{0.588} &\textbf{0.533}\\
\midrule
\rowcolor{Gray} \multicolumn{17}{c}{\textbf{SD-Prompts}~\cite{Santana2022StableDiffusionPrompts}} \\
\midrule
Clean    & 0.944 & 0.998 & 1.000 & 1.000 & 1.000 & 1.000 & 0.990 & 0.819 & 1.000& 0.991 & 0.994 & 1.000 & 0.970 & 1.000 & 1.000 & 0.972\\ 
\midrule
Bright.  & 0.471 & 0.748 & 0.742 & 0.972 & 0.955 & 0.999 & 0.815 & 0.557 & 0.537 & 0.823 & 0.899 & 0.949 & 0.922 & 0.992 & 1.000 & 0.835\\
Cont.    & 0.894 & 0.999 & 1.000 & 1.000 & 1.000 & 1.000 & 0.982 & 0.516 & 0.459 & 0.963 & 0.967 & 0.999 & 0.928 & 0.988 & 0.983 & 0.850\\
JPEG     & 0.466 & 0.979 & 0.990 & 1.000 & 0.999 & 0.989 & 0.904 & 0.506 & 0.610 & 0.810 & 0.769 & 0.990 & 0.936 & 1.000& 1.000 & 0.828\\
Blur     & 0.912 & 0.999 & 1.000 & 1.000 & 1.000 & 0.981 & 0.982 & 0.685 & 0.999 & 0.988 & 0.888 & 0.998 & 0.910 & 0.994 & 0.980 & 0.930\\
Noise    & 0.423 & 0.903 & 0.850 & 0.988 & 0.992 & 1.000 & 0.859 & 0.822 & 0.998 & 0.961 & 0.742 & 0.992 &  0.819& 0.960 & 0.948 & 0.905\\
BM3D     & 0.802 & 1.000 & 1.000 & 1.000 & 1.000 & 0.876 & 0.946 & 0.530 & 0.859 & 0.915 & 0.809 & 1.000 &  0.864& 1.000& 0.971 & 0.869\\
\midrule
Center Crop     & 0.469 & 0.994 & 1.000 & 1.000 & 1.000 & 0.860 & 0.887 & 0.723 & 0.743 & 0.980 & 0.983 & 1.000 & 0.872 & 0.893 & 0.920 & 0.889\\
Random Crop     & 0.749 & 0.996 & 1.000 & 1.000 & 1.000 & 0.799 & 0.924 & 0.794 & 0.860 & 0.983 & 0.990 & 1.000 & 0.749 & 0.891& 0.912 & 0.897\\
\midrule
VAE-B~\cite{balle2018variational}    & 0.509 & 0.940 & 0.983 & 0.996 & 0.996 & 0.849 & 0.879 & 0.513 & 0.659 & 0.572 & 0.677 & 0.976 & 0.820 & 0.853 & 0.862 & 0.742\\
VAE-C~\cite{cheng2020learned}    & 0.514 & 0.975 & 0.987 & 1.000 & 0.999 & 0.881 & 0.893 & 0.512 & 0.620 & 0.535 & 0.671 & 0.994 & 0.829 & 0.881& 0.900 &  0.743\\
Regen.~\cite{zhao2024invisible}   & 0.543 & 0.958 & 0.999 & 1.000 & 1.000 & 0.857 & 0.893 & 0.509 & 0.623 & 0.567 & 0.493 & 0.999 & 0.803 &0.852 & 0.890 & 0.717\\
Rinse    & 0.498 & 0.872 & 0.967 & 0.983 & 0.991 & 0.783 & 0.849 & 0.493 & 0.547 & 0.528 & 0.511 & 0.951 & 0.751 & 0.839 & 0.872 & 0.687\\
\midrule
CtrlGen+~\cite{liu2024image}      & 0.076 & 0.283 & 0.768 & 1.000 & 1.000 & 0.127& 0.542 & 0.494 & \textbf{0.471} & 0.548 & 0.591 & 1.000 & 0.655 &\textbf{0.562} & 0.813 & 0.642\\
UnMarker~\cite{kassis2025unmarker}     & 0.028 & 0.084 & \textbf{0.036} & 0.284 & 0.024 & 0.043& 0.083 &0.485 &  0.507  & 0.548 & \textbf{0.511}& 0.579&0.523 &0.658 & 0.604 & 0.552\\
\midrule
\texttt{RAVEN}   & \textbf{0.025} & \textbf{0.062} & 0.038 & \textbf{0.024} & \textbf{0.020} &\textbf{0.018} & \textbf{0.031} & \textbf{0.471} & 0.504 &  \textbf{0.477}  & 0.523&\textbf{0.562} &\textbf{0.480} & 0.587 & \textbf{0.600} & \textbf{0.526}\\
\bottomrule
\end{tabular}
\end{adjustbox}
\label{tab:identify_transposed}
\end{table*}

\subsubsection{View-Guided Correspondence Attention}

While latent viewpoint modulation introduces the spatial perturbation 
needed to disrupt watermark alignment, naively denoising the transformed 
latent $\tilde{z}_\tau$ is insufficient. Without guidance, the diffusion 
model treats $\tilde{z}_\tau$ as an independent image, causing appearance 
drift, color shifts, and loss of fine details. To preserve perceptual 
correspondence with the original scene, we modify the attention mechanism 
within the diffusion UNet.
In standard diffusion models $\epsilon_\theta(z_t, t)$, self-attention operates only on the current noisy latent $\text{SelfAttn}(Q, K, V) = \text{softmax}\!\left(\frac{QK^\top}{\sqrt{d}}\right)V,$
where $Q = W_Q z_t$, $K = W_K z_t$, and $V = W_V z_t$. This restricts 
all interactions to a single frame and provides no mechanism to anchor 
denoising to the original content after a viewpoint transformation.

We introduce view-guided attention that couples the transformed latent 
$\tilde{z}_t$ with a reference latent $z_t^{\text{ref}}$ denoised in 
parallel from the untransformed inverted latent $z_\tau$. At each 
timestep $t$, queries are computed from the transformed latent while 
keys and values come from the reference:
\begin{equation}
\begin{small}
\text{ViewAttn}(Q, K, V) = \text{softmax}\!\left(\frac{(W_Q \tilde{z}_t)(W_K z_t^{\text{ref}})^\top}{\sqrt{d}}\right)W_V z_t^{\text{ref}}, 
\end{small}
\end{equation}

This mechanism forces each position in $\tilde{z}_t$ to attend to 
semantically corresponding features in $z_t^{\text{ref}}$, preserving 
appearance, texture, and object identity throughout denoising. Critically, 
the attention operates in learned feature space rather than pixel 
coordinates—each query matches to regions in the reference based on 
semantic similarity, naturally accommodating the spatial misalignment 
introduced by viewpoint modulation. This allows the diffusion model to 
reconstruct a perceptually coherent alternative view while the watermark, 
which depends on precise pixel-level spatial correlation, remains 
disrupted. The view-guided correspondence enforced at each denoising 
timestep prevents semantic drift, ensuring geometric perturbations 
selectively remove the watermark without degrading scene fidelity.

\subsubsection{Color and Contrast Transfer}

While view-guided correspondence attention preserves structural and semantic fidelity, minor color shifts or contrast inconsistencies may still arise due to stochastic denoising. We found that a simple color and contrast transfer in the CIELAB space effectively restores perceptual quality without reintroducing the watermark. Let $x_{\text{opt}} = \mathcal{D}(\tilde{z}_0)$ denote the denoised image obtained from the final latent $\tilde{z}_0$ after the diffusion process, with CIELAB components $\{L_{\text{opt}}, a_{\text{opt}}, b_{\text{opt}}\}$, and let $\{L_w, a_w, b_w\}$ denote the corresponding components of the original watermarked image $x_w$. For \textbf{color transfer}, we retain the luminance from the optimized image and adopt the chrominance from the watermarked one: $x_c = \mathcal{F}_{\text{RGB}}(L_{\text{opt}}, a_w, b_w)$, where $\mathcal{F}_{\text{RGB}}$ converts from CIELAB to RGB space. To further improve visual consistency, \textbf{contrast transfer} aligns the luminance statistics between $x_c$ and $x_w$ as $L_{\text{final}} = \frac{\sigma_w}{\sigma_c}(L_c - \mu_c) + \mu_w$, where $\mu_c, \sigma_c$ and $\mu_w, \sigma_w$ are the mean and standard deviation of the luminance channels of $x_c$ and $x_w$, respectively. The final image is reconstructed as $\tilde{x} = x_{\text{final}} = \mathcal{F}_{\text{RGB}}(L_{\text{final}}, a_w, b_w)$. Although these steps are simple, they effectively correct color artifacts and enhance local contrast, complementing our diffusion-based stages.

\section{Experiments}

\subsection{Experiment Settings}

\noindent
\textbf{Model Setup and Datasets.} 
We evaluate \texttt{RAVEN} on three prompt datasets: 5,000 captions from MS-COCO-2017~\cite{lin2014microsoft}, 1,001 prompts from DiffusionDB 2M~\cite{wang2023diffusiondb}, and 8,192 prompts from Stable-Diffusion-Prompts~\cite{Santana2022StableDiffusionPrompts}. Images are generated at 512×512 resolution using Stable Diffusion v2-1~\cite{rombach2022high} with CFG scale 7.5 and 50 DDIM steps. Watermark detection experiments use 1,000 image pairs to measure TPR@1\%FPR.

\definecolor{Gray}{gray}{0.93}
\begin{SCtable*}[\sidecaptionrelwidth][t]
\small
\renewcommand{\arraystretch}{0.93}
\caption{\small Average image-quality metrics (FID and CLIP-Text Score) across 16 watermarking schemes and three datasets. Lower FID reflects improved perceptual fidelity, while higher CLIP-Text scores indicate better semantic consistency with the prompt. Best results are highlighted in bold.}
\centering
\begin{adjustbox}{width=0.75\textwidth}
\begin{tabular}{lcccccccccccccc}
\toprule
\rowcolor{Gray}\multirow{2}{*}{Metric} & \multicolumn{5}{c}{\textbf{Signal Processing}} & 
\multicolumn{5}{c}{\textbf{Regeneration}} & 
\multicolumn{2}{c}{\textbf{Advanced}} & 
\multirow{2}{*}{Ours} \\
\cmidrule(lr){2-6} \cmidrule(lr){7-11} \cmidrule(lr){12-13}
\rowcolor{Gray}& Bright. & Blur & Noise & R.C. & C.C. & VAE-B & VAE-C & Regen. & Rinse2x & Rinse4x & CtrlGen+ & UnMarker & \\
\midrule
\rowcolor[rgb]{0.9,0.95,1.0}\multicolumn{14}{c}{\textbf{MS-COCO}~\cite{lin2014microsoft}} \\
\midrule
FID $\downarrow$    & 90.97 & 45.07 & 112.83 & 64.29  & 110.92 & 44.91 & 42.12 & 42.36 & 47.92 & 51.18 & 40.96 & 49.85 & \textbf{40.18} \\
CLIP $\uparrow$     & 0.312 & 0.325 & 0.307 & 0.312 & 0.320 & 0.324 & 0.328 & 0.327 & 0.319 & 0.302 & 0.323 & 0.316 & \textbf{0.328} \\
\midrule
\rowcolor[rgb]{0.9,0.95,1.0}\multicolumn{14}{c}{\textbf{SD-Prompts}~\cite{Santana2022StableDiffusionPrompts}} \\
\midrule
FID $\downarrow$    & 121.34 & 53.29 & 134.29 & 77.32 & 130.60 & 53.21 & 51.99 & 49.78 & 52.06 & 58.83 & \textbf{48.95} & 55.48 & 49.47 \\
CLIP $\uparrow$     &  0.338 & 0.352 & 0.330 & 0.342 & 0.346 & 0.352 & 0.354 & 0.358 & 0.356 & 0.331 & 0.364 & 0.353 & \textbf{0.364} \\
\midrule
\rowcolor[rgb]{0.9,0.95,1.0}\multicolumn{14}{c}{\textbf{DiffusionDB}~\cite{wang2023diffusiondb}} \\
\midrule
FID $\downarrow$    & 107.56 &53.89  & 129.11 & 78.39 & 121.38 & 52.71 & 48.10 & 47.83 & 51.96 & 55.89 & 47.33 & 50.69 & \textbf{47.11} \\
CLIP $\uparrow$     & 0.316 & 0.345 & 0.319 & 0.347 & 0.339 & 0.348 & 0.350 & 0.348 & 0.341 & 0.339 & 0.348 & 0.344 & \textbf{0.350} \\
\bottomrule
\end{tabular}
%\vspace{-1em}
\end{adjustbox}
\label{tab:quality_metrics}
\end{SCtable*}
\begin{figure*}[t]
    \centering
    \includegraphics[width=\linewidth,trim={1em 4em 0.5em 2em},clip]{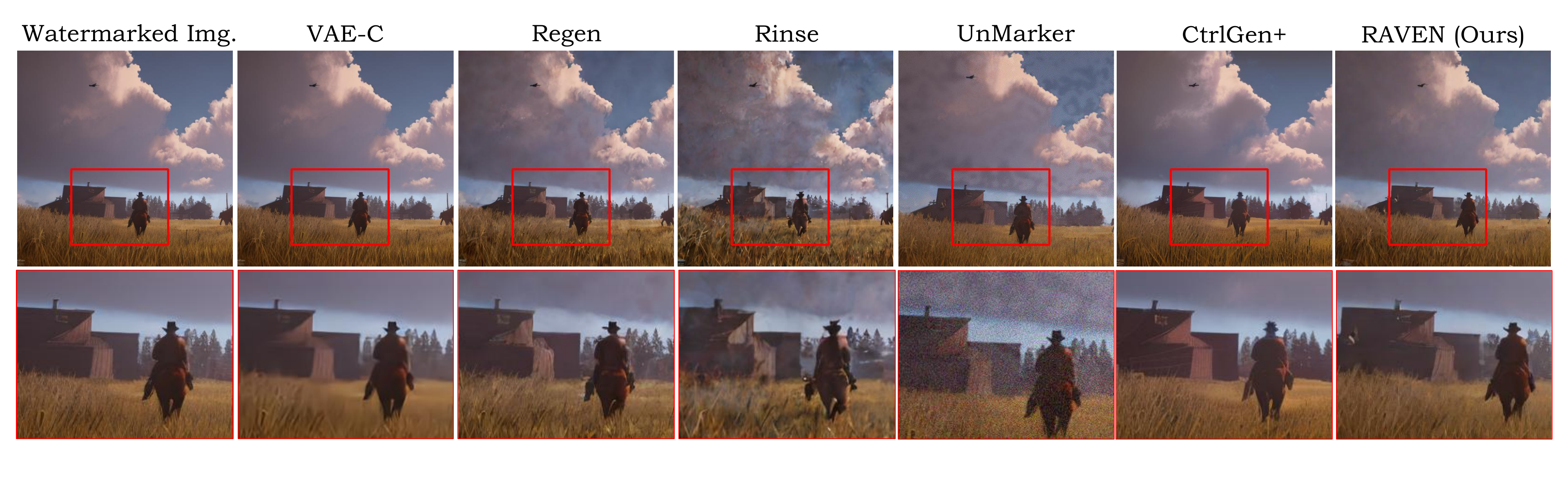}
    \caption{\small \textbf{Qualitative comparison of watermark removal methods}. The top row shows full images, while the bottom row displays zoomed-in regions (red boxes) for detailed inspection. VAE-C~\cite{cheng2020learned} introduces excessive blurring that degrades fine details. Regen~\cite{zhao2024invisible} produces visible artifacts due to high noise injection required for watermark removal. Rinse exhibits unnatural color shifts and loss of photorealism, a consequence of performing multiple regeneration passes. UnMarker~\cite{liu2024image} leaves noisy residual artifacts that compromise visual quality. CtrlGen+~\cite{liu2024image} produces overly stylized outputs that deviate from natural appearance. In contrast, \texttt{RAVEN} preserves fine-grained details, natural textures, and photorealistic appearance.  Note that the zoomed regions for UnMarker~\cite{liu2024image} and \texttt{RAVEN} differ slightly from other methods due to cropping layers and camera translation transformations, respectively. More results are provided in suppl.}
    \label{fig:main_qualitative}
    \vspace{-1.5em}
\end{figure*}

\noindent \textbf{Image Watermark Methods.} 
We evaluate \texttt{RAVEN} on 14 watermarking methods using their official implementations, including both \textit{post-hoc} and \textit{in-generation} schemes: DwtDct~\cite{cox2008digital}, DwtDctSvd~\cite{cox2008digital}, RivaGAN~\cite{zhang2019robust}, StegaStamp~\cite{tancik2020stegastamp}, StableSignature~\cite{fernandez2023stable}, Tree-Ring~\cite{wen2023tree}, Zodiac~\cite{zhang2024robust}, HSTR~\cite{lee2025semantic}, RingID~\cite{ci2024ringid}, HSQR~\cite{lee2025semantic}, Gaussian Shading~\cite{yang2024gaussian}, ROBIN~\cite{huang2024robin},  TrustMark~\cite{bui2023trustmark}, and VINE~\cite{lu2024robust}. Most methods are designed with robustness as a core objective, providing a rigorous benchmark for evaluating \texttt{RAVEN}'s removal efficacy.

\noindent \textbf{Image Watermark Removal Baselines.}
We evaluate \texttt{RAVEN} against 14 watermark removal attacks across three categories. 
\textit{(i) Classical Signal Processing:} brightness adjustment, contrast modification, JPEG compression, Gaussian blur, additive Gaussian noise, and BM3D denoising~\cite{dabov2007image}. These pixel-space operations serve as widely used, low-cost baselines. 
\textit{(ii) Regeneration Attacks:} Regen~\cite{zhao2024invisible} regenerates watermarked images via diffusion-based noising and denoising. Rinse iteratively applies Regen multiple times for improved removal. We also test two VAE-based compression models at quality level 3: VAE-B~\cite{balle2018variational} and VAE-C~\cite{cheng2020learned}.
\textit{(iii) Advanced Methods:} CtrlGen+~\cite{liu2024image} performs controllable regeneration from clean-noise initialization, while UnMarker~\cite{kassis2025unmarker} is a universal optimization-driven attack for heterogeneous watermarking schemes. 
These baselines provide a rigorous benchmark for evaluating \texttt{RAVEN}'s robustness and fidelity. We exclude methods requiring surrogate models, such as IRA~\cite{muller2025black} or adversarial attacks~\cite{an2024waves}, as their performance depends on surrogate quality whereas \texttt{RAVEN} operates in a black-box setting.

\noindent \textbf{Implementation Details.} 
We use the Image-to-Image Stable Diffusion Refiner with frozen parameters to preserve its generative behavior. Classifier-free guidance scale is set to 2.5, with 50-step DDIM inversion and image-to-image strength of 0.15. Since the original prompt is unknown in our threat model, we use empty text prompts for both inversion and reconstruction. Latent viewpoint modulation applies diagonal translations randomly sampled from [24, 32] or [-32, -24] pixels along each axis. All experiments run on a single NVIDIA A100 GPU with deterministic inference using fixed random seeds for reproducibility.

\noindent
\textbf{Evaluation Metrics.} 
Following recent work~\cite{liu2024image, kassis2025unmarker}, we report TPR@1\%FPR for verification-based semantic watermarking and Bit Accuracy for bitstream-based schemes. We assess generation quality using FID~\cite{heusel2017gans} between attacked and original watermarked images. Semantic alignment is measured via CLIP score using OpenCLIP-ViT/G~\cite{cherti2023reproducible}, quantifying preservation of prompt alignment.

\subsection{Main Results}

\noindent \textbf{Watermark Removal Performance.}
Tab.~\ref{tab:identify_transposed} presents watermark removal results across 16 methods on MS-COCO and SD-Prompts, reporting TPR@1\%FPR for semantic methods (lower is better) and Bit Accuracy for bitstream methods (0.5 indicates randomization). Classical signal processing attacks show limited effectiveness, while regeneration-based methods achieve moderate suppression. Advanced methods demonstrate stronger removal, with UnMarker~\cite{kassis2025unmarker} achieving the best baseline performance at 0.078 average TPR on MS-COCO. \texttt{RAVEN} outperforms all baselines, achieving 0.026 average TPR in MS-COCO—representing more than 60\% improvement over the closest baseline. Consistent superior performance is observed across both datasets and bitstream-based schemes. DiffusionDB results show similar trends (see suppl.).

\noindent \textbf{Quality Preservation.}
Tab.~\ref{tab:quality_metrics} reports FID and CLIP-Text scores averaged across 16 watermarking schemes on three datasets. Signal processing attacks introduce substantial distortion, with FID ranging from 45.07 to 134.29. Regeneration-based methods achieve better fidelity, with VAE-C and Regen showing FID around 42 to 52 across datasets. \texttt{RAVEN} achieves the best or near-best scores across all datasets: FID of 40.18 on MS-COCO, 49.47 on SD-Prompts, and 47.11 on DiffusionDB, while maintaining the highest CLIP scores of 0.328, 0.364, and 0.350 respectively. These results demonstrate that \texttt{RAVEN} effectively balances watermark removal with superior perceptual fidelity and semantic consistency.

\noindent \textbf{Qualitative Results.} 
Fig.~\ref{fig:main_qualitative} compares watermark removal attacks outputs. Pixel-space baselines struggle with quality preservation: VAE-C~\cite{cheng2020learned} excessively blurs details, Regen~\cite{zhao2024invisible} introduces regeneration artifacts, and Rinse produces unnatural textures. UnMarker~\cite{kassis2025unmarker} leaves residual noise patterns, while CtrlGen+~\cite{liu2024image} yields overly stylized, over-smoothed outputs. In contrast, \texttt{RAVEN} preserves fine-grained details and natural textures. Additional qualitative results are provided in suppl.

\subsection{Ablations}
\begin{figure}[t]
    \centering
    \includegraphics[width=\linewidth]{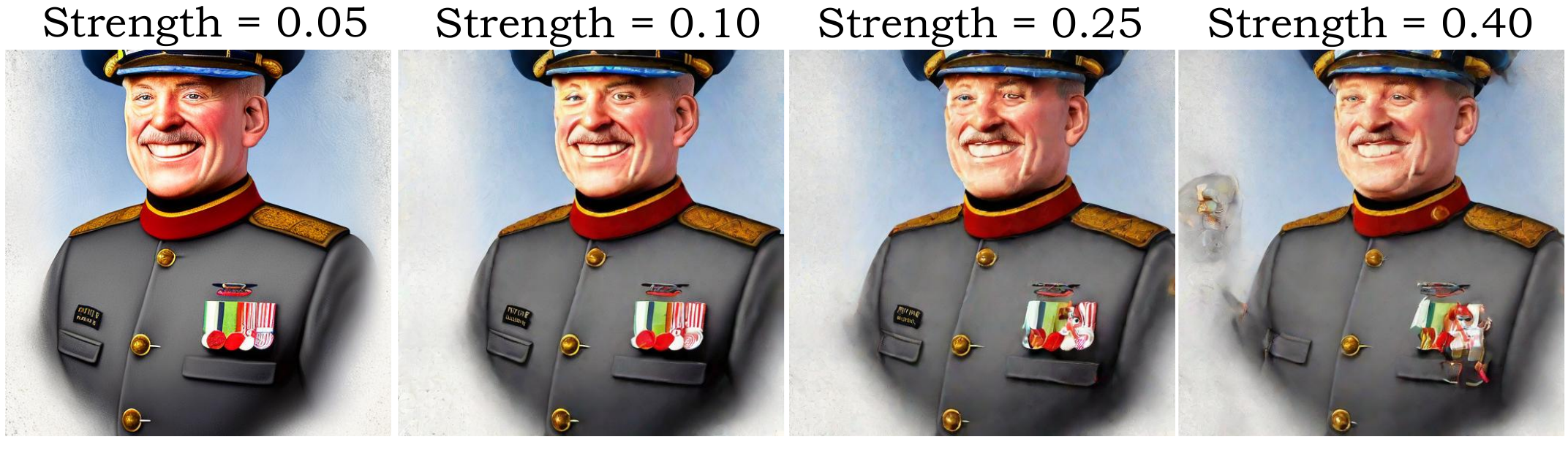}
\caption{\textbf{Effect of strength parameter $s$ on watermark removal.} Increasing $s$ enhances watermark suppression but degrades visual quality. Low values ($s = 0.05$) preserve quality but may retain watermarks, while high values introduce artifacts.}
    \label{fig:strength_ablation}
\end{figure}

\begin{table}[t]
\centering
\caption{\textbf{Effect of strength parameter on visual quality.}}
\label{tab:strength_ablation}
\small
\resizebox{0.48\textwidth}{!}{%
\begin{tabular}{lcccccc}
\toprule
Strength ($s$) & 0.10 & 0.15 & 0.20 & 0.25 & 0.35 & 0.45 \\
\midrule
FID $\downarrow$ & 62.68 & 65.50 & 69.01 & 75.43 & 79.62 & 85.10 \\
\bottomrule
\end{tabular}%
}
\end{table}
\begin{figure}[t]
    \centering
    \includegraphics[width=\linewidth]{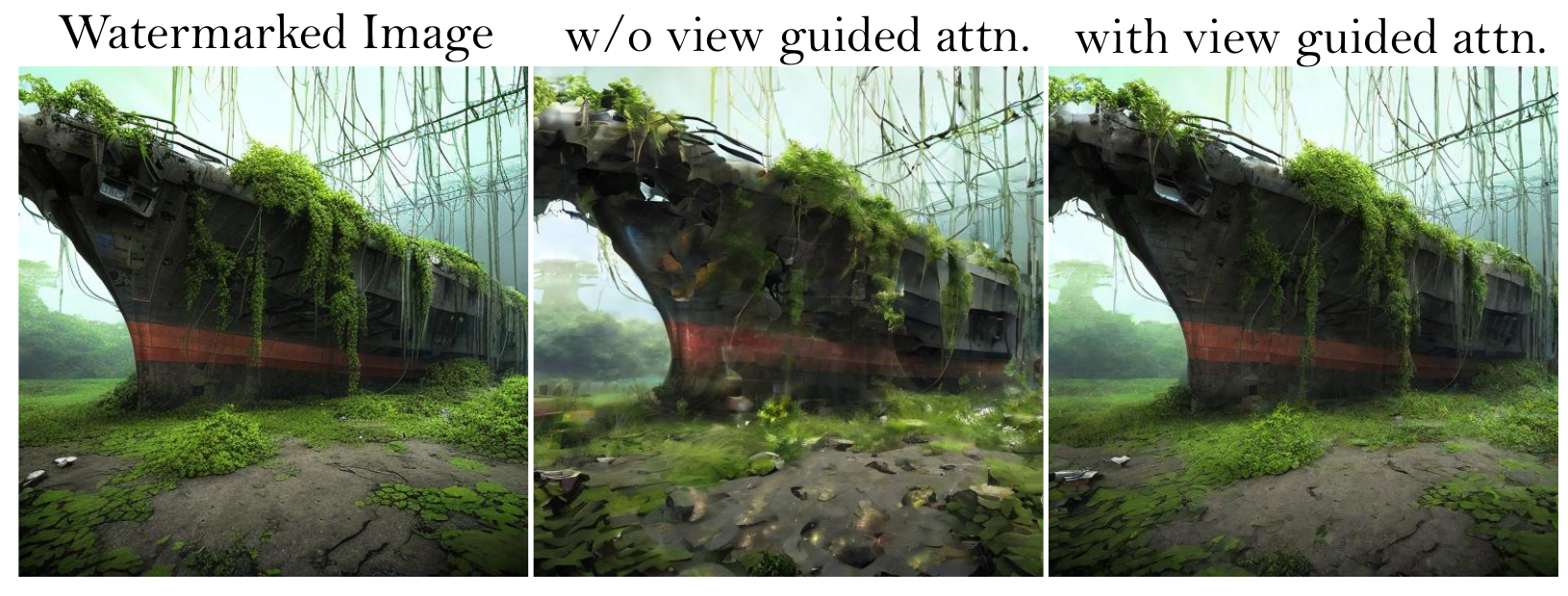}
    \caption{\textbf{Effect of view-guided correspondence attention.} Without correspondence attention (middle), latent viewpoint modulation causes severe structural distortions. With view-guided attention (right), \texttt{RAVEN} preserves fine-grained details, textures, and structural consistency while effectively removing watermarks.}
    \label{fig:attention_ablation}
\end{figure}

\begin{table}[t]
\centering
\scriptsize
\setlength{\tabcolsep}{4.5pt}
\caption{\textbf{Model-agnostic watermark removal.} \texttt{RAVEN} achieves consistent watermark suppression across different Stable Diffusion Image-to-Image backbones, demonstrating generalizability without model-specific tuning.}
\label{tab:model_agnostic}
\resizebox{\columnwidth}{!}{
\begin{tabular}{lcccc}
\toprule
Model & TPR@1\%FPR $\downarrow$ & Bit Acc. & FID $\downarrow$ & CLIP $\uparrow$ \\
\midrule
Img2Img v2.1 & 0.026 & 0.523 & 40.18 & 0.328 \\
Img2Img v2.0 & 0.026 & 0.519 & 42.61 & 0.328 \\
Img2Img v1.5 & 0.029 & 0.547 & 44.93 & 0.325 \\
\bottomrule
\end{tabular}
}
\vspace{-3pt}
\end{table}

\begin{figure}[t]
    \centering
    \includegraphics[width=1\linewidth,trim={0em 0em 0em 0em},clip]{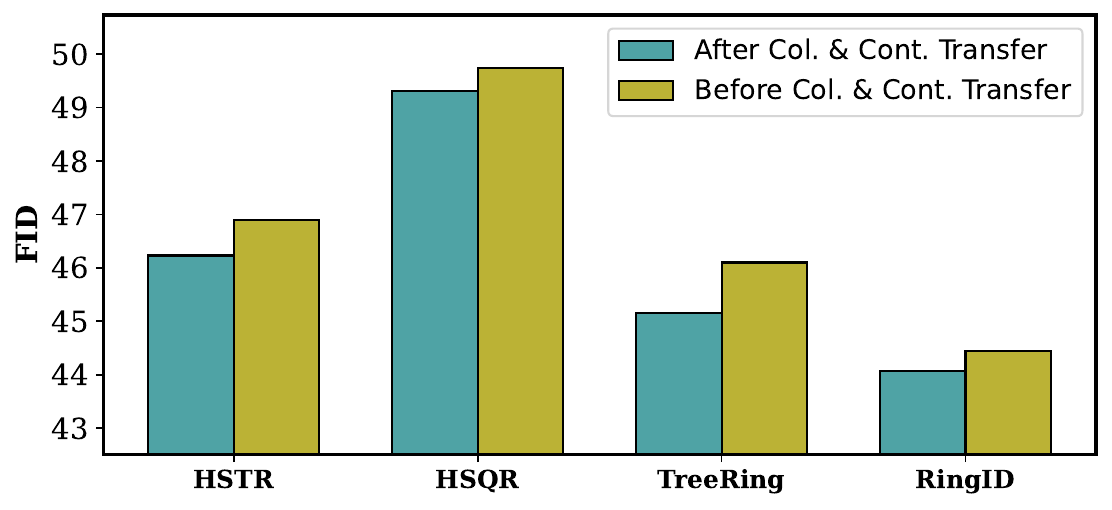}
\caption{\textbf{Effect of color and contrast transfer.} FID comparison across four watermarking methods before and after applying color and contrast transfer in CIELAB space. The post-processing step consistently improves FID.}
\label{fig:color_contrast_ablation}
\end{figure}

\noindent \textbf{Effect of Strength Parameter.} 
The strength parameter $s$ controls noise injection during partial diffusion inversion, trading off watermark removal and visual quality. Fig.~\ref{fig:strength_ablation} and Tab.~\ref{tab:strength_ablation} show that low $s$ preserves quality but may retain watermarks, while high $s$ improves suppression but introduces artifacts and degrades FID.

\noindent \textbf{Impact of View-Guided Correspondence Attention.}
Fig.~\ref{fig:attention_ablation} demonstrates that removing correspondence attention causes structural distortion. Incorporating view-guided attention preserves semantic consistency and fine textures without introducing watermark, confirming its crucial role in maintaining perceptual quality.

\noindent \textbf{Impact of Color and Contrast Transfer.} Fig.~\ref{fig:color_contrast_ablation} shows that applying color and contrast transfer in CIELAB space consistently improves FID of \texttt{RAVEN} outputs across all watermarking methods, correcting color shifts and contrast inconsistencies without restoring watermark signals.

\noindent \textbf{Model-Agnostic Generalization.}
Tab.~\ref{tab:model_agnostic} shows that \texttt{RAVEN} achieves consistent watermark suppression across Stable Diffusion v1.5, v2.0, and v2.1 backbones without model-specific tuning. All versions maintain strong performance with TPR below 0.03 and comparable visual quality, demonstrating that our view synthesis approach generalizes across diffusion architectures.

\section{Conclusion}

We introduced a novel attack vector against invisible watermarks by framing removal as view synthesis. Our zero-shot diffusion-based framework applies controlled viewpoint shifts in latent space combined with view-guided correspondence attention to remove watermarks while preserving semantic content. Without requiring detector access or watermark knowledge, \texttt{RAVEN} achieves state-of-the-art suppression across 14 watermarking schemes and 15 attack baselines. This work exposes a critical vulnerability to semantic-preserving geometric transformations, motivating the design of more resilient watermarking mechanisms.

{
    \small
    \bibliographystyle{ieeenat_fullname}
    \bibliography{main}
}

% CVPR 2026 Paper Template; see https://github.com/cvpr-org/author-kit

\twocolumn[
\begin{@twocolumnfalse}
\section*{\centering \Large{{Supplementary Material} -- \texttt{RAVEN}: Erasing Invisible Watermarks via Novel View Synthesis}}
\vspace{2em}
\end{@twocolumnfalse}
]

This supplementary material provides additional experimental evidence to support the effectiveness and robustness of \texttt{RAVEN} for invisible watermark removal. We extend our evaluation by presenting \textbf{comprehensive quantitative results on the DiffusionDB dataset}~\cite{wang2023diffusiondb}, covering a wide range of watermarking schemes and attack strategies, and analyzing verification performance in terms of TPR@1\%FPR and Bit Accuracy. We also include \textbf{qualitative comparisons across challenging visual scenarios} to illustrate how different watermark removal methods impact perceptual quality, structural consistency, and texture preservation. In addition to effectiveness, we report the \textbf{computational efficiency} of \texttt{RAVEN}, highlighting its practical advantage over other strong baselines. Overall, these supplementary results further demonstrate \texttt{RAVEN}'s ability to achieve strong watermark suppression while maintaining high visual fidelity and semantic coherence, complementing the findings reported in the main paper.

\section{Quantitative Results on DiffusionDB}

Table~\ref{tab:diffusiondb_results} reports watermark removal performance on the DiffusionDB dataset~\cite{wang2023diffusiondb}, a large and diverse collection of real-world text-to-image prompts. We evaluate \texttt{RAVEN} on 1,001 randomly sampled prompts using the same protocol as the main paper, generating images at 512$\times$512 resolution with Stable Diffusion v2-1, CFG scale 7.5, and 50 DDIM steps. The evaluation spans the same 15 watermarking schemes, including both semantic in-generation and post-hoc bitstream methods, and compares \texttt{RAVEN} against 14 baseline attacks covering classical signal processing, regeneration-based approaches, and advanced learned methods such as CtrlGen+~\cite{liu2024image} and UnMarker~\cite{kassis2025unmarker}, ensuring a fair and consistent comparison across all paradigms.

\subsection{Key Findings}

Results on DiffusionDB~\cite{wang2023diffusiondb} reinforce the trends observed in the main paper, with \texttt{RAVEN} achieving the lowest average TPR@1\%FPR of \textbf{0.029} for semantic watermarking, substantially outperforming the strongest baseline, UnMarker, and demonstrating consistently superior suppression across all six methods, with particularly notable gains for challenging schemes such as RingID and ROBIN. For bitstream-based watermarking, \texttt{RAVEN} attains an average bit accuracy of \textbf{0.531}, close to the ideal randomization range and comparable to UnMarker, while preserving noticeably better perceptual quality. Performance remains highly stable across datasets, with semantic TPR and bit accuracy varying by less than 0.005 between DiffusionDB~\cite{wang2023diffusiondb}, MS-COCO~\cite{lin2014microsoft}, and SD-Prompts~\cite{Santana2022StableDiffusionPrompts}, confirming that \texttt{RAVEN} leverages fundamental properties of diffusion latent representations and watermark embedding mechanisms rather than dataset-specific prompt characteristics.

\section{Additional Qualitative Results}

\begin{figure*}[t]
    \centering
    \includegraphics[width=0.95\linewidth,trim={2em 1em 0.5em 2em},clip]{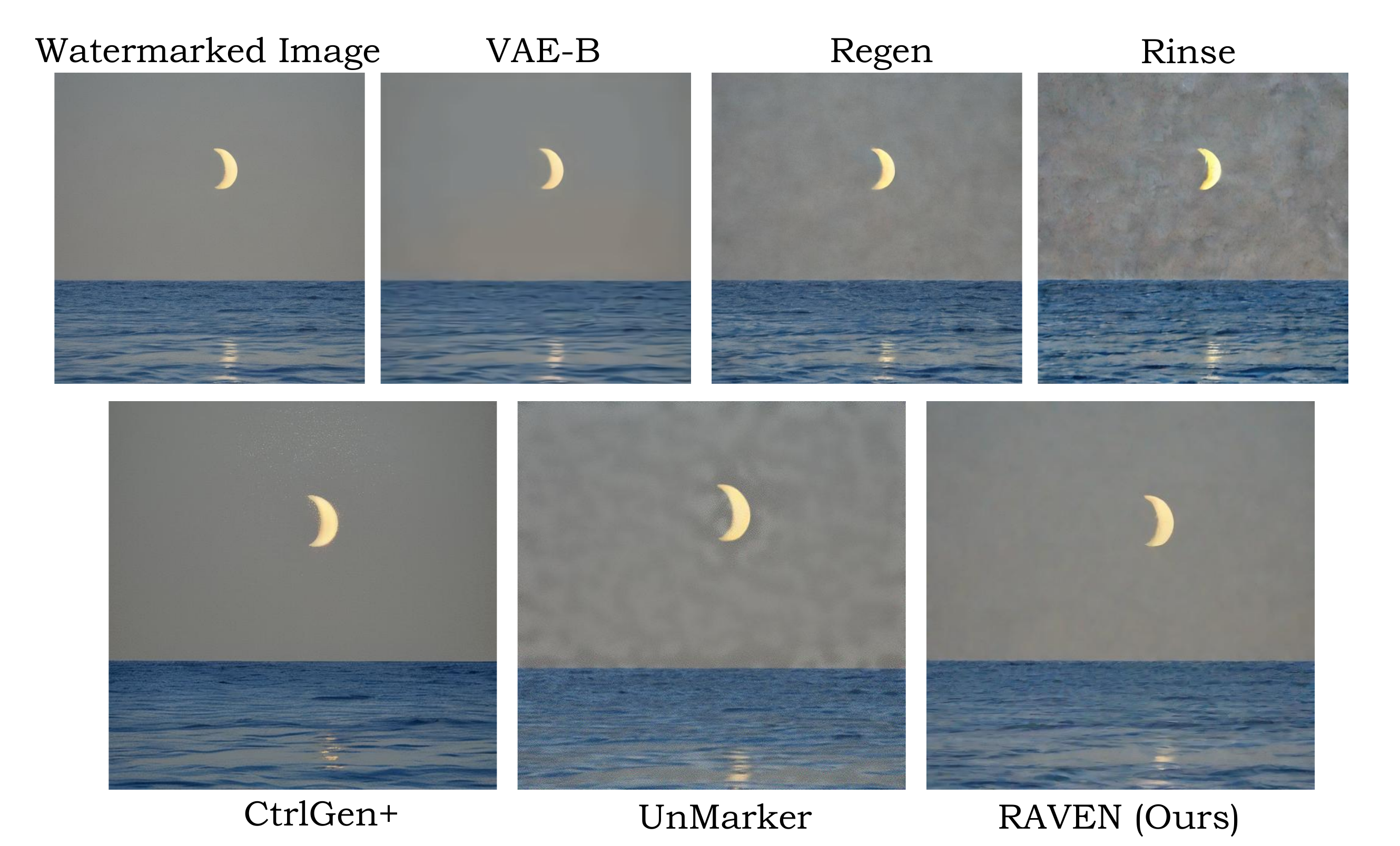}
    \caption{\small \textbf{Qualitative comparison of watermark removal methods}.  VAE-B~\cite{cheng2020learned} introduces excessive blurring that degrades fine details. Regen~\cite{zhao2024invisible} produces visible artifacts due to high noise injection required for watermark removal. Rinse exhibits unnatural color shifts and loss of photorealism, a consequence of performing multiple regeneration passes. UnMarker~\cite{liu2024image} leaves noisy residual artifacts that compromise visual quality. CtrlGen+~\cite{liu2024image} produces overly stylized outputs that deviate from natural appearance. In contrast, \texttt{RAVEN} preserves fine-grained details, natural textures, and photorealistic appearance.  Note that the images for UnMarker~\cite{liu2024image} and \texttt{RAVEN} differ slightly from other methods due to cropping layers and camera translation transformations, respectively.}
    \label{fig:main_qualitative1}
    \vspace{-1.5em}
\end{figure*}

\begin{figure*}[t]
    \centering
    \includegraphics[width=0.84\linewidth,trim={2em 1em 0.5em 2em},clip]{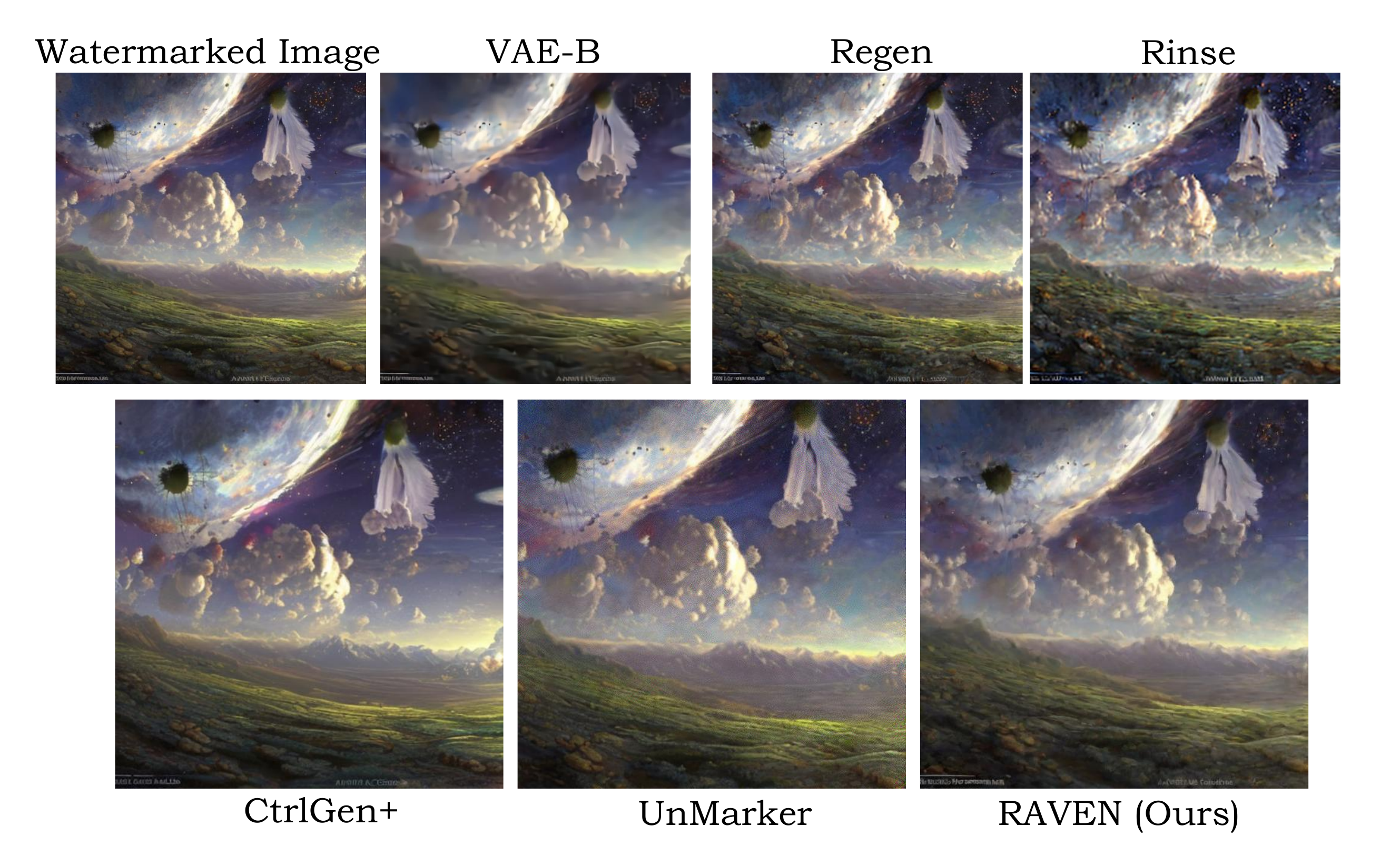}
    \caption{\small \textbf{Qualitative comparison of watermark removal methods}.  VAE-B~\cite{cheng2020learned} introduces excessive blurring that degrades fine details. Regen~\cite{zhao2024invisible} produces visible artifacts due to high noise injection required for watermark removal. Rinse exhibits unnatural color shifts and loss of photorealism, a consequence of performing multiple regeneration passes. UnMarker~\cite{liu2024image} leaves noisy residual artifacts that compromise visual quality. CtrlGen+~\cite{liu2024image} produces overly stylized outputs that deviate from natural appearance. In contrast, \texttt{RAVEN} preserves fine-grained details, natural textures, and photorealistic appearance.  Note that the images for UnMarker~\cite{liu2024image} and \texttt{RAVEN} differ slightly from other methods due to cropping layers and camera translation transformations, respectively.}
    \label{fig:main_qualitative2}
    \vspace{-1.5em}
\end{figure*}

\begin{figure*}[t]
    \centering
    \includegraphics[width=0.84\linewidth,trim={2em 1em 0.5em 2em},clip]{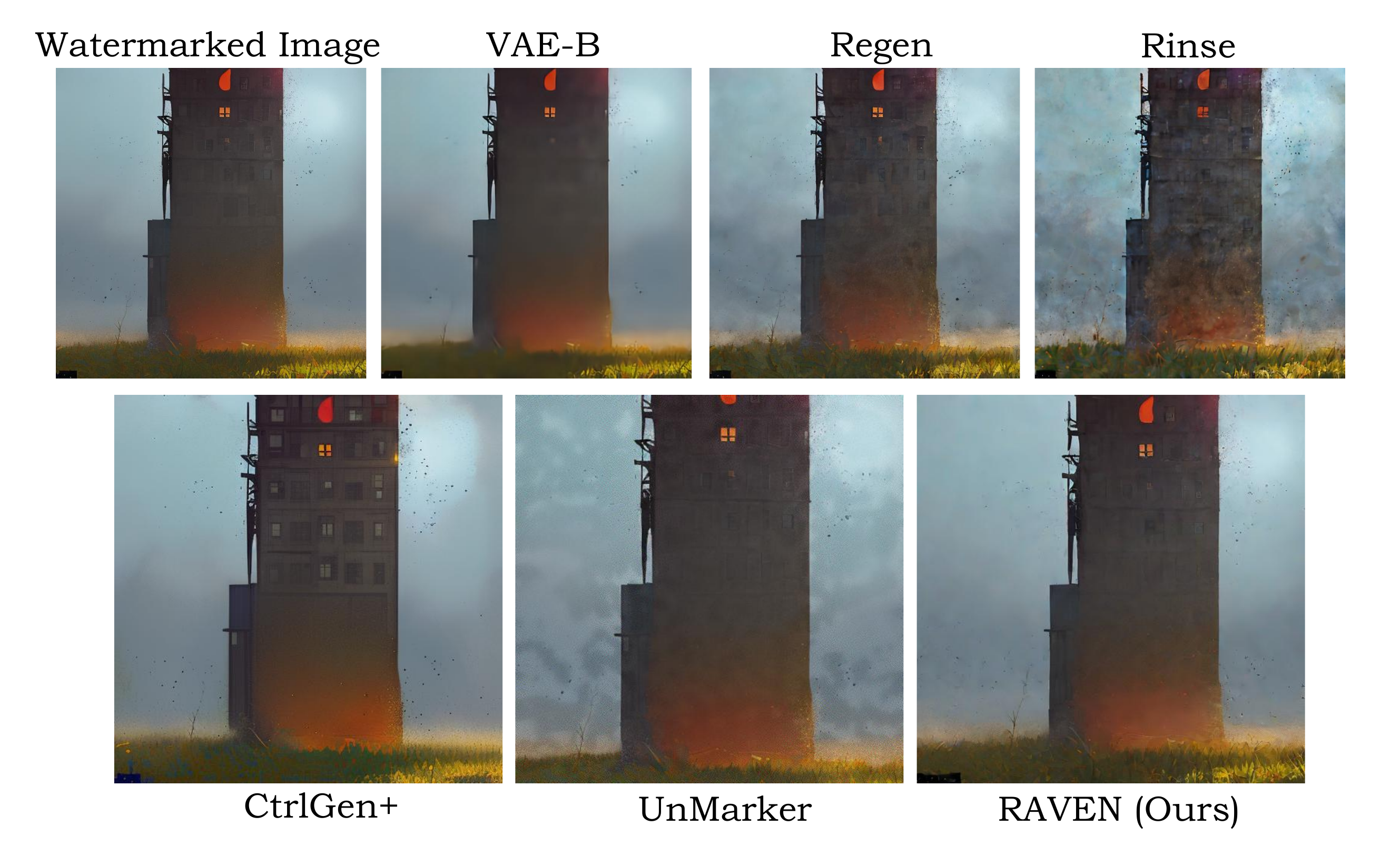}
    \caption{\small \textbf{Qualitative comparison of watermark removal methods}.  VAE-B~\cite{cheng2020learned} introduces excessive blurring that degrades fine details. Regen~\cite{zhao2024invisible} produces visible artifacts due to high noise injection required for watermark removal. Rinse exhibits unnatural color shifts and loss of photorealism, a consequence of performing multiple regeneration passes. UnMarker~\cite{liu2024image} leaves noisy residual artifacts that compromise visual quality. CtrlGen+~\cite{liu2024image} produces overly stylized outputs that deviate from natural appearance. In contrast, \texttt{RAVEN} preserves fine-grained details, natural textures, and photorealistic appearance.  Note that the images for UnMarker~\cite{liu2024image} and \texttt{RAVEN} differ slightly from other methods due to cropping layers and camera translation transformations, respectively.}
    \label{fig:main_qualitative3}
    \vspace{-1.5em}
\end{figure*}

Figures~\ref{fig:main_qualitative1}-\ref{fig:main_qualitative2} present qualitative comparisons of watermark removal methods across diverse visual scenes, highlighting clear differences in perceptual quality and structural preservation. VAE-B~\cite{balle2018variational} consistently introduces excessive smoothing that suppresses fine details and weakens texture sharpness, while Regen~\cite{zhao2024invisible} generates visible high-frequency artifacts due to the aggressive noise injection required for watermark removal. Rinse often leads to unnatural color shifts and reduced photorealism as a result of repeated regeneration passes. UnMarker~\cite{kassis2025unmarker} suppresses watermark signals more effectively than conventional priors but leaves residual noise patterns and minor structural inconsistencies. Notably, as evident in the Figure~\ref{fig:main_qualitative3}, CtrlGen+~\cite{liu2024image} significantly alters the underlying scene layout, introducing new architectural or geometric elements and generating alternative structures that deviate from the original semantic configuration rather than preserving it. In contrast, \texttt{RAVEN} maintains faithful scene geometry, consistent structure, and natural texture distribution while successfully eliminating watermark traces, resulting in visually coherent and photorealistic outputs that closely align with the original watermarked content.

\section{Computational Efficiency}

Among advanced methods that achieve strong watermark suppression, \texttt{RAVEN} offers a favorable balance between effectiveness and efficiency. UnMarker~\cite{kassis2025unmarker} requires approximately 5 minutes per image on an NVIDIA A100 GPU due to its iterative optimization process, while CtrlGen+\cite{liu2024image} requires multi-GPU training infrastructure with 8 NVIDIA A100 GPUs to achieve its reported performance. In contrast, \texttt{RAVEN} operates in a zero-shot manner without additional training and processes each image in approximately 6 seconds on an A100 40GB GPU. This efficiency is enabled by leveraging frozen pretrained diffusion models with lightweight attention modifications, avoiding costly per-image optimization or large-scale training requirements. While signal processing attacks are faster, their limited effectiveness against modern watermarking schemes makes them less relevant in practical scenarios.

\begin{table*}[t]
\centering
\scriptsize
\caption{Verification performance of different watermarking methods under various attacks on DiffusionDB dataset. TPR@1\%FPR is reported for in-generation semantic watermarking methods (TreeRing to ROBIN), where lower values indicate better attack performance. Bit Accuracy is reported for post-hoc bitstream-based methods (DwtDct to VINE), where values near 0.5 indicate successful watermark randomization. \texttt{RAVEN} achieves the lowest detection rates across both categories, demonstrating superior removal efficacy while maintaining visual quality.}
\begin{adjustbox}{width=\textwidth}
\begin{tabular}{l|>{\columncolor{myblue!50}}c>{\columncolor{myblue!50}}c>{\columncolor{myblue!50}}c>{\columncolor{myblue!50}}c>{\columncolor{myblue!50}}c>{\columncolor{myblue!50}}c|c|>{\columncolor{mygreen!50}}c>{\columncolor{mygreen!50}}c>{\columncolor{mygreen!50}}c>{\columncolor{mygreen!50}}c>{\columncolor{mygreen!50}}c>{\columncolor{mygreen!50}}c>{\columncolor{mygreen!50}}c>{\columncolor{mygreen!50}}c|c}
\toprule
\small \rotatebox{70}{Attack} & \rotatebox{70}{TreeRing~\cite{wen2023tree}} & \rotatebox{70}{Zodiac~\cite{zhang2024robust}} & \rotatebox{70}{HSTR~\cite{lee2025semantic}} & \rotatebox{70}{RingID~\cite{ci2024ringid}} & \rotatebox{70}{\shortstack{HSQR~\cite{lee2025semantic}\\\textcolor{blue}{\scriptsize }}} & \rotatebox{70}{ROBIN~\cite{huang2024robin}} & \rotatebox{70}{Avg.} & \rotatebox{70}{DwtDct~\cite{cox2008digital}} & \rotatebox{70}{DwtDctSvd~\cite{cox2008digital}} & \rotatebox{70}{RivaGAN~\cite{zhang2019robust}} & \rotatebox{70}{Stable Sign.~\cite{fernandez2023stable}} & \rotatebox{70}{Gauss Shad.~\cite{yang2024gaussian}} & \rotatebox{70}{TrustMark~\cite{bui2023trustmark}} & \rotatebox{70}{Stega St.~\cite{tancik2020stegastamp}} & \rotatebox{70}{VINE~\cite{lu2024robust}} & \rotatebox{70}{Avg.}\\
\midrule
\rowcolor{Gray} \multicolumn{17}{c}{\textbf{DiffusionDB}~\cite{wang2023diffusiondb}} \\
\midrule
Bright.  & 0.487 & 0.752 & 0.792 & 0.989 & 0.977 & 0.987 & 0.831 & 0.563 & 0.588 & 0.839 & 0.890 & 0.954 & 0.913 & 0.989 & 0.995 & 0.841\\
Cont.    & 0.889 & 0.988 & 0.996 & 1.000 & 1.000 & 0.991 & 0.977 & 0.515 & 0.463 & 0.960 & 0.967 & 0.999 & 0.922 & 0.994 & 0.989 & 0.851\\
JPEG     & 0.434 & 0.933 & 0.981 & 1.000 & 0.999 & 0.984 & 0.905 & 0.509 & 0.593 & 0.790 & 0.787 & 0.990 & 0.913 & 1.000 & 1.000 & 0.848\\
Blur     & 0.904 & 0.988 & 0.996 & 1.000 & 1.000 & 0.985 & 0.979 & 0.672 & 0.997 & 0.985 & 0.889 & 0.999 & 0.931 & 0.988 & 0.970 & 0.929\\
Noise    & 0.392 & 0.834 & 0.792 & 0.963 & 0.974 & 0.895 & 0.808 & 0.829 & 0.995 & 0.937 & 0.726 & 0.992 & 0.793 & 0.964 & 0.972 & 0.889\\
BM3D     & 0.799 & 0.984 & 0.991 & 1.000 & 0.999 & 0.869 & 0.940 & 0.526 & 0.830 & 0.893 & 0.813 & 0.998 & 0.880 & 0.994 & 0.997 & 0.879\\
\midrule
Center Crop     & 0.499 & 0.971 & 1.000 & 1.000 & 1.000 & 0.887 & 0.893 & 0.723 & 0.742 & 0.974 & 0.981 & 0.999 & 0.883 & 0.917 & 0.945 & 0.895\\
Random Crop     & 0.715 & 0.985 & 1.000 & 1.000 & 1.000 & 0.795 & 0.916 & 0.801 & 0.860 & 0.979 & 0.986 & 1.000 & 0.739 & 0.888 & 0.910 & 0.906\\
\midrule
VAE-B    & 0.454 & 0.911 & 0.968 & 0.995 & 0.997 & 0.824 & 0.858 & 0.513 & 0.658 & 0.553 & 0.690 & 0.978 & 0.821 & 0.869 & 0.906 & 0.748\\
VAE-C    & 0.503 & 0.926 & 0.969 & 0.999 & 0.999 & 0.847 & 0.874 & 0.514 & 0.608 & 0.518 & 0.687 & 0.989 & 0.833 & 0.875 & 0.909 & 0.742\\
Regen.   & 0.454 & 0.903 & 0.989 & 1.000 & 1.000 & 0.865 & 0.869 & 0.512 & 0.621 & 0.556 & 0.496 & 0.998 & 0.796 & 0.857 & 0.886 & 0.715\\
Rinse    & 0.445 & 0.861 & 0.975 & 0.990 & 0.996 & 0.799 & 0.844 & 0.501 & 0.559 & 0.531 & 0.507 & 0.964 & 0.738 & 0.825 & 0.858 & 0.685\\
\midrule
CtrlGen+   & 0.084 & 0.300 & 0.767 & 1.000 & 1.000 & 0.311 & 0.577 & 0.523 & 0.509 & 0.516 & 0.581 & 1.000 & 0.676 & 0.562 & 0.881 & 0.656\\
UnMarker     & 0.031 & 0.090 & 0.034 & 0.265 & 0.021 & 0.042 & 0.081 & 0.489 & 0.517 & 0.538 & 0.510 & 0.597 & 0.542 & 0.651 & 0.623 & 0.559\\
\midrule
\texttt{RAVEN}    & {0.023} & {0.070} & {0.028} & {0.022} & {0.020} & {0.015} & \textbf{0.029} & {0.515} & {0.491} & {0.502} & {0.522} & {0.550} & {0.485} & {0.583} & {0.597} & \textbf{0.531}\\
\bottomrule
\end{tabular}
\end{adjustbox}
\label{tab:diffusiondb_results}
\end{table*}

\end{document}